\documentclass[letterpaper, 10 pt, journal, twoside]{IEEEtran}

\IEEEoverridecommandlockouts
\usepackage{cite}
\usepackage{amsmath,amssymb,amsfonts}
\usepackage{algorithmic}
\usepackage{graphicx}
\usepackage{textcomp}
\usepackage{xcolor}
\usepackage{balance}
\usepackage{comment}
\usepackage[ruled,vlined]{algorithm2e}
\usepackage{caption}
\usepackage{array}
\usepackage{subcaption}
\usepackage{arydshln}
\usepackage{tensor}
\newcommand{\eg}{\emph{e.g.},}
\newcommand{\ie}{\emph{i.e.},}

\usepackage{hyperref}
\usepackage{float}
\usepackage{rotating}
\usepackage{nicematrix}  %
\usepackage{mathtools}
\usepackage{amsmath}
\usepackage{pifont}
\usepackage{cuted}
\usepackage{dblfloatfix}

\usepackage{titlesec}
\titlespacing*{\section}{0pt}{.5em}{0pt}

\definecolor{LightCyan}{rgb}{0.88,1,0.88}
\definecolor{linear_color}{RGB}{220,223,240}
\definecolor{gray_bbox_color}{RGB}{243,243,244}

\definecolor{rebuttal}{rgb}{0,0,1}

\def\eqref#1{Eq.~(\ref{#1})}

\newcommand{\cmark}{\ding{51}}%

\def\BibTeX{{\rm B\kern-.05em{\sc i\kern-.025em b}\kern-.08em
    T\kern-.1667em\lower.7ex\hbox{E}\kern-.125emX}}
\begin{document}

\title{\LARGE \bf GeoAdapt: Self-Supervised Test-Time Adaptation in LiDAR Place Recognition Using Geometric Priors
}
\author{Joshua Knights$^{1,2}$, Stephen Hausler$^{1}$, Sridha Sridharan$^{2}$, Clinton Fookes$^{2}$, Peyman Moghadam$^{1,2}$
\thanks{Manuscript received: August, 2, 2023; Revised October, 31, 2023; Accepted November, 20, 2023.}
\thanks{This paper was recommended for publication by Editor Cesar Cadena Lerma upon evaluation of the Associate Editor and Reviewers' comments.}
\thanks{
$^1$ CSIRO Robotics and Autonomous Systems, DATA61, CSIRO, 
Australia. 
E-mails: {\tt\footnotesize \emph{firstname.lastname}@csiro.au}}
\thanks{
$^{2}$ Signal Processing, AI and Vision Technologies (SAIVT),
Queensland University of Technology (QUT), Brisbane, Australia.
E-mails: {\tt\footnotesize \emph\{s.sridharan, c.fookes, peyman.moghadam\}@qut.edu.au}
}
\thanks{Digital Object Identifier (DOI): see top of this page.}

} 

\markboth{IEEE Robotics and Automation Letters. Preprint Version. Accepted November, 2023}
{Knights \MakeLowercase{\textit{et al.}}: GeoAdapt}

\bstctlcite{IEEEexample:BSTcontrol}

\maketitle

\begin{strip}
    \centering
    \vspace{-4.1cm}
    \includegraphics[width=\textwidth]{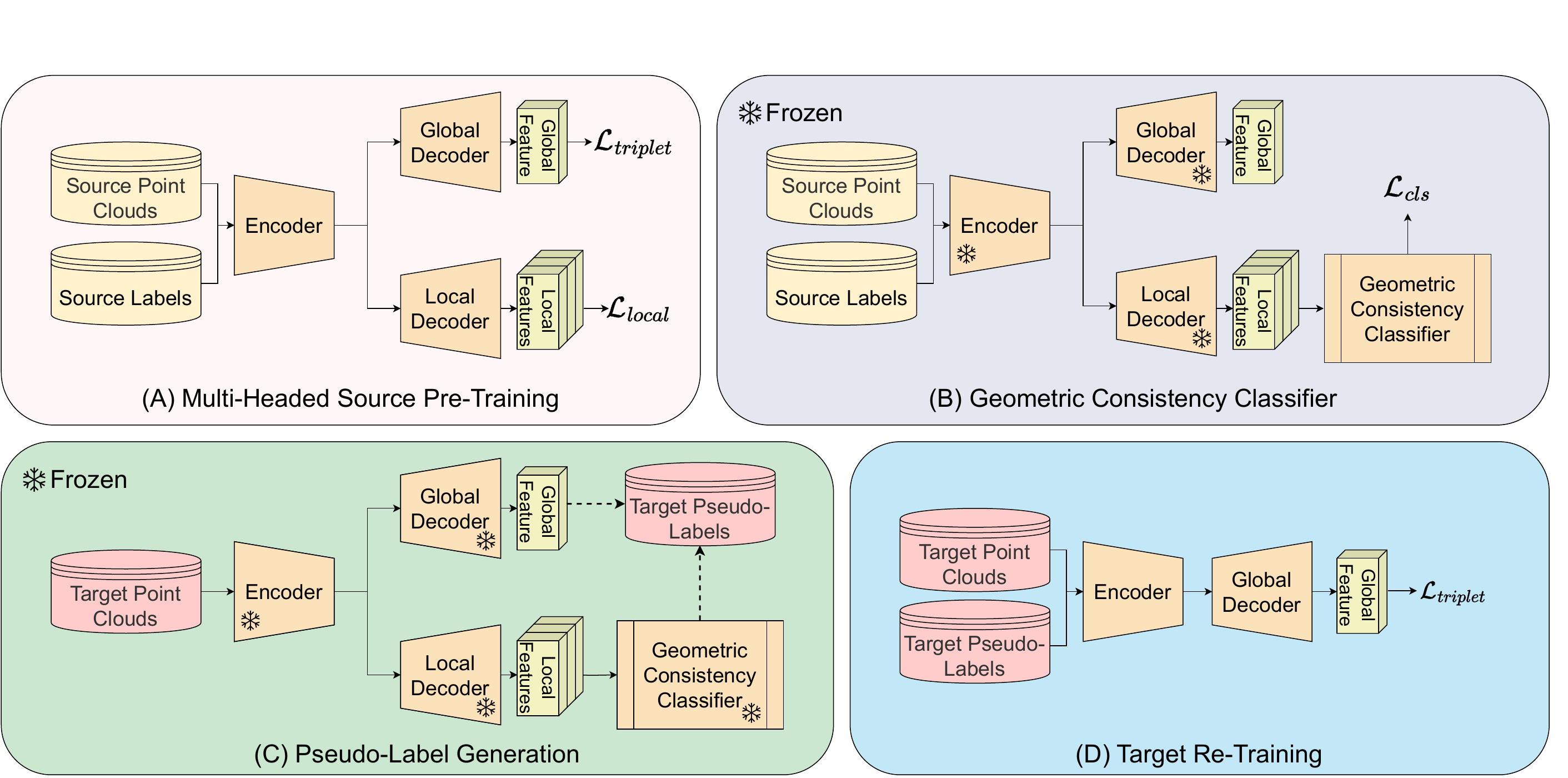}
    \captionof{figure}{Overview of \textit{GeoAdapt}, our proposed Test-Time Adaptation approach.  \textit{GeoAdapt} adapts a model pre-trained on a source domain (A) by using a novel auxiliary classification head, which uses geometric consistency as a prior to classify a pair of clouds as positively or negatively associated (B).  We then use the auxiliary head to generate pseudo-labels for pairs of point clouds on the target domain, using the $L_2$ similarity of the global features to guide which pairs to generate pseudo-labels for (C).  These pseudo-labels are used to adapt the model's parameters to the target domain to improve test-time performance (D).} 
    
    \label{fig:overview}
    \vspace{-0.5cm}
\end{strip}

\vspace{2mm}
\begin{abstract}
LiDAR place recognition approaches based on deep learning suffer from significant performance degradation when there is a shift between the distribution of training and test datasets, often requiring re-training the networks to achieve peak performance.  However, obtaining accurate ground truth data for new training data can be prohibitively expensive, especially in complex or GPS-deprived environments.  To address this issue we propose \textit{GeoAdapt}, which introduces a novel auxiliary classification head to generate pseudo-labels for re-training on unseen environments in a self-supervised manner.  \textit{GeoAdapt} uses geometric consistency as a prior to improve the robustness of our generated pseudo-labels against domain shift, improving the performance and reliability of our Test-Time Adaptation approach.  Comprehensive experiments show that \textit{GeoAdapt} significantly boosts place recognition performance across moderate to severe domain shifts, and is competitive with fully supervised test-time adaptation approaches.  Our code is available at \href{https://github.com/csiro-robotics/GeoAdapt}{https://github.com/csiro-robotics/GeoAdapt}.

\end{abstract}


\section{Introduction}
\label{sec:intro}
\IEEEPARstart{P}{lace} recognition, the task of recognizing previously visited locations when traversing an environment, is a vital component of many applications in embodied intelligence and is essential for reliable loop closures in simultaneous localization and mapping (SLAM) or global relocalisation when deployed in a GPS-denied environment.  The current state-of-the-art for place recognition is dominated by deep learning-based methods \cite{Komorowski_2021_WACV,komorowski2022improving,vid2022logg3d,wiesmann2022kppr,hui2021pptnet}, which achieve remarkable performance when training and test data share a similar distribution. However, the performance of these learning-based approaches degrades significantly in the presence of domain shifts between the training and test data.  {While using a probabilistic approach can be sufficient to overcome this gap in the case of minor domain shifts \cite{bhutta2022so},  variations in the sensor, adversarial weather conditions, or major environmental shifts (\eg{} urban to natural) can pose significant challenges to the generalisability of LiDAR place recognition models, impacting their ability to be deployed over a wider range of platforms and environments \cite{ramezani2023deep}.}%

A straightforward solution to addressing this problem is to refine a model by re-training using additional labelled samples on test environment.  However, ground truth for training place recognition models is generally obtained using GPS or SLAM, leading to difficulties when re-training on environments which are either GPS-denied or sufficiently large and complex enough to introduce failures into the SLAM solution.  In addition, the source data initially used to train the model may not be available at test-time due to privacy or accessibility reasons.  Ideally, the model should be adapted to the test data in a self-supervised manner, using only the pre-trained model parameters from the target domain.  This setting is known as Test-Time Adaptation (TTA).

Test-Time Adaptation describes the task of refining a model pre-trained on a \textit{source} dataset using an unlabelled \textit{target} dataset, accommodating for the distribution shift between the source and target domains. 
Many existing TTA approaches \cite{Liang2020DoWR,Shin_2022_CVPR,Zhang2021AuxAdaptSA,litrico_2023_CVPR} employ pseudo-labelling, where predictions from the source pre-trained model are used to guide re-training on the target dataset.  However, pseudo-labels generated by the pre-trained model usually contain significant noise due to the domain gap between the source and target datasets, resulting in poor adaptation when they are used naively for re-training.  {In addition, TTA approaches for tasks such as segmentation and object detection are generally inapplicable to LiDAR place recognition due to fundamental differences in the nature of the ground truth required; for tasks in the former category, the ground truth denotes properties of the individual training samples such as semantic class or object locations, while for LiDAR place recognition the ground truth instead denotes the relationship between point clouds (\ie{} are they \textit{positively} or \textit{negatively} associated).}

In this work, we propose a novel TTA approach for LiDAR place recognition which we call \textit{GeoAdapt}, shown in Figure \ref{fig:overview}.  {\textit{GeoAdapt} formulates the task of pseudo-label generation for place recognition as a classification problem, introducing a novel auxiliary classification head which classifies pairs of point clouds as positively or negatively associated in a registration-free manner for re-training without access to the target ground-truth.}  Specifically, the auxiliary classification head uses local feature matching to propose inlier correspondences between a pair of input point clouds and estimates the confidence of these correspondences using the geometric consistency of the input point clouds as a prior.  We observe that positively associated point clouds are often geometrically consistent, while it is highly unlikely that the correspondences proposed for negatively associated point clouds will exhibit such geometrically consistent behaviour.  By classifying the predicted confidence of our proposed inliers we significantly reduce the likelihood of noise in our generated pseudo-labels, resulting in reliable TTA without requiring any access to data or supervision from the source and target domains respectively.  %

We comprehensively benchmark \textit{GeoAdapt} using five large-scale, public LiDAR datasets (two source and three target datasets), and demonstrate that our proposed approach consistently improves place recognition performance on out-of-distribution target domains.  Notably, we show that our approach is competitive with re-training using ground truth supervision on the target domain, demonstrating the strength of using geometric consistency as the basis of self-supervision when generating labels for training.  Our contributions are as follows:
\begin{itemize}
    \item {We introduce \textit{GeoAdapt}, a novel Test-Time Adaptation approach which allows for re-training a model on unseen target environments without access to any ground truth supervision.  Our approach employs a novel auxiliary classification head which uses geometric consistency as a prior to generate pseudo-labels in a registration-free manner which are robust against domain shifts, improving the effectiveness of our adaptation to target environments.}
    \item We validate our method using five large-scale, public LiDAR datasets (2 source and 3 target datasets), and show our approach improves performance on unseen target domains by a large margin and achieves competitive performance with fully supervised adaptation.
\end{itemize}

\label{sec:rel_work}

\section{Related Work}

\subsection{LiDAR Place Recognition}
\label{sec:lpr}
LiDAR Place Recognition (LPR) is formulated as a retrieval problem, where different methods encode a point cloud into a compact global descriptor which can be used to query a database of previously visited places.  Existing approaches can be broadly categorised into handcrafted~\cite{kim2018scan,scancontextppTRO}, hybrid \cite{vidanapathirana2020locus,xu2021disco}, or fully end-to-end deep learning \cite{Komorowski_2021_WACV,komorowski2022improving,vid2022logg3d,wiesmann2022kppr,hui2021pptnet} approaches.  Deep learning approaches in particular demonstrate remarkable performance when the training and test data share a similar distribution, but can have their generalisability on unseen data suffer significantly in the presence of a domain shift \cite{knights2022incloud}.  A key commonality of deep learning based approaches is the use of metric loss functions such as the triplet \cite{Komorowski_2021_WACV,komorowski2022improving}, quadruplet \cite{uy2018pointnetvlad,vid2022logg3d} or contrastive \cite{wiesmann2022kppr} loss, which are reliant on the selection of hard positive or negative examples using ground truth sensor pose information to form training tuples.  Obtaining accurate sensor pose information is often done using either GPS or SLAM, with both approaches introducing challenges.  Target environments may be GPS-denied, and even when available, GPS-based ground truth can result in false or noisy positive and negative examples when sensor occlusion or field of view are not taken into consideration~\cite{Brachmann2021OnTL}. Using SLAM to generate the ground truth suffers from the drawback of requiring accurate loop closures, which necessitates high-performing place recognition solutions beforehand.

\subsection{Test-Time Adaptation}

Test-Time Adaptation (TTA) aims to adapt a model pre-trained on a labelled source domain to improve performance on an unseen and unlabelled target domain, where there is a significant domain shift between the source and target.  Related tasks include domain generalization \cite{Kim_2023_CVPR,xiao20233d} which aims to train models using only the source data to perform broadly well across a variety of conditions, and online TTA \cite{Wang2022ContinualTD,Shin_2022_CVPR}, which aims to adapt the model online from a stream of data during evaluation.  A common approach \cite{Liang2020DoWR,Shin_2022_CVPR,Zhang2021AuxAdaptSA,litrico_2023_CVPR} to TTA is to leverage predictions from the pre-trained model to predict pseudo-labels on the target domain, which are used for re-training the model to close the domain gap.  These pseudo-labels are generally noisy as a result of the domain gap between source and target domains, with recent approaches using auxiliary heads \cite{Zhang2021AuxAdaptSA} or uncertainty estimation \cite{litrico_2023_CVPR} to refine the pseudo-labels for re-training.  In this work, we propose an auxiliary classification head which employs geometric priors relating to the geometric consistency of matching point clouds to produce robust and reliable pseudo-labels, allowing for effective target adaptation even under moderate to severe domain gaps.

Works exploring domain adaptation and generalization in 3D perception have focused primarily on the tasks of semantic segmentation \cite{xiao20233d,Shin_2022_CVPR,Kim_2023_CVPR,saltori2022cosmix,Ryu_2023_CVPR} and object detection \cite{Wei2022LiDARDB,Lehner20213DVFieldAA,wang2023ssda3d,yang2021st3d}, with a particular focus on the domain gaps induced by either sim-to-real \cite{DODA2022,Huch2023QuantifyingTL} or changing LiDAR sensors \cite{Ryu_2023_CVPR} between the source and target.
However, the problem of adaptation for LiDAR place recognition has gone largely unexplored in the literature to date.  InCloud \cite{knights2022incloud} re-trains the model on new domains to address the challenge of continual learning, but requires access to the target ground truth to do so.  Continual SLAM \cite{Vodisch2022ContinualSB} adapts the depth and odometry prediction of a SLAM system deployed in an unfamiliar environment, but freezes its loop closure (\ie{} place recognition) components during deployment.  SGV \cite{vidanapathirana2023sgv} and Uncertainty-LPR~\cite{mason2022uncertainty} improve test-time performance by re-ranking the top retrieved candidates and filtering out uncertain examples respectively, but will still struggle when the top retrieved candidates are not well calibrated on the target domains.  The closest existing work to ours is \cite{Lajoie2022SelfSupervisedDC}, which performs domain adaptation for visual place recognition but relies on SLAM to identify positive and negative examples for training. Specifically, they used pose graph optimization to calculate which pairs of images are positives or negatives, based on an initial set of potential place recognition matching pairs. However, as they note in their work, pose graph optimization can fail even with just a single incorrect loop closure. Therefore, during severe domain shifts, it is more likely that SLAM-based TTA will fail to find accurate positive and negative examples. Our approach does not suffer from this limitation, and can successfully adapt to severe domain shifts such as from urban to natural environments. Our proposed approach formulates pseudo-label generation for place recognition as a classification task and avoids using SLAM or any other pose estimation techniques as a prior, allowing for simple and effective adaptation of a model between source and target domains.

\section{Problem Formulation} 
Presume we have a place recognition model pre-trained on a \textit{source} dataset $\mathcal{S} = \left\{\left(\mathcal{P}_1,t_1\right), \left(\mathcal{P}_2,t_2\right), \cdots \left(\mathcal{P}_N,t_N\right)\right\}$, where $\mathcal{P}\in\mathbb{R}^{n\times 3}$ denotes a point cloud and $t$ denotes the ground truth sensor pose for that point cloud.  Now also presume we have a second \textit{target} dataset $\mathcal{T}=\left\{\mathcal{P}_1,\mathcal{P}_2,\cdots \mathcal{P}_N\right\}$ that contains point clouds with no accompanying ground truth.  When there are significant shifts in the distribution of $\mathcal{S}$ and $\mathcal{T}$ due to differences in sensor, environment, or other factors, the model pre-trained on $\mathcal{S}$ will suffer from degraded place recognition performance on $\mathcal{T}$ as a result of the aforementioned domain shift.  The aim of TTA is to adapt the parameters of a model pre-trained on $\mathcal{S}$ to improve performance on $\mathcal{T}$ without having any access to source data or ground truth supervision at test time.

\section{GeoAdapt}
In this section we propose \textit{GeoAdapt}, a TTA approach for LiDAR place recognition which introduces a novel auxiliary classification head which outputs a pair-likelihood estimation for any given pair of input point clouds.
TTA using \textit{GeoAdapt} is split into four different steps.  First, we pre-train a model on $\mathcal{S}$ to produce both a global descriptor and dense local features from different encoders.  Secondly, we freeze the model and train our proposed Geometric Consistency Classifier using the frozen local features, which uses geometric consistency as a prior in order to produce pair-likelihood scores which are highly robust against domain shift.  Thirdly, we use the pair-likelihood scores produced by the auxiliary classifier to generate pseudo-labels for $\mathcal{T}$, and finally we use these pseudo-labels to re-train the model on $\mathcal{T}$ without requiring source data or target supervision.  Figure \ref{fig:overview} provides an overview of our model design and adaptation pipeline.

\subsection{Multi-Headed Source Pre-Training}

The backbone of our approach is a multi-headed model architecture consisting of a global feature head and an auxiliary classification head.  The global feature head consists of a shared encoder and a shallow decoder which outputs a global feature vector for an input point cloud, while the auxiliary classification head consists of the shared encoder, a local feature decoder which outputs dense point features for the input point cloud, and our proposed Geometric Consistency Classifier.  Before training the Geometric Consistency Classifier, we pre-train the shared encoder and both the local and global feature decoders on the source dataset.  To train the global feature we use the triplet loss which is defined as:

\begin{figure}[t]
    \centering
    \includegraphics[width=\columnwidth]{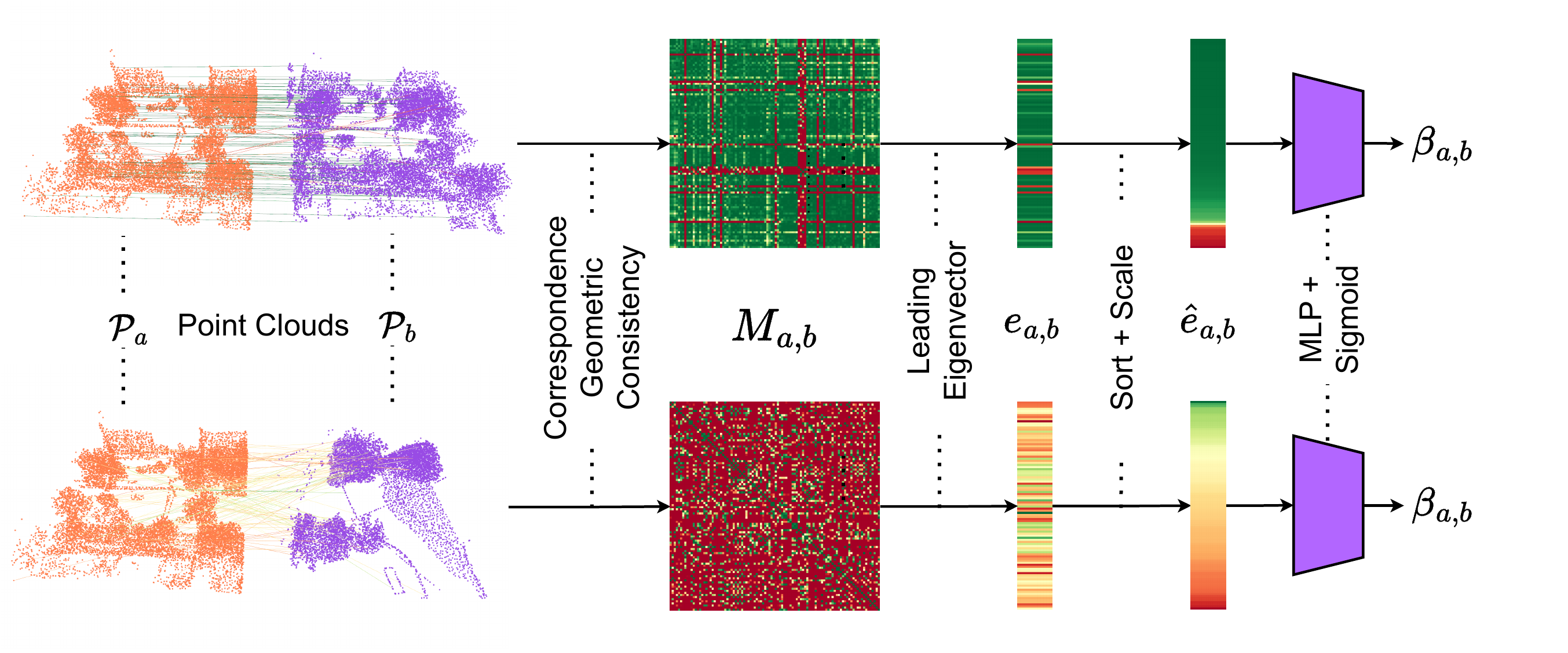}
    \caption{%
    Geometric Consistency Classifier.  Positively associated point clouds (top) produce correspondences which are highly geometrically consistent, while negatively associated point clouds (bottom) produce correspondences with little to no consistency.  The geometric consistency matrix $M_{a,b}$ is used to estimate the inlier confidence, and then to predict the pair-likelihood score for a given pair of input point clouds.}
    \label{fig:spatial_consistency}
    \vspace{-5mm}
\end{figure}

\begin{equation}
    \mathcal{L}_{triplet} = \left[\left\|g_{anc}-g_{pos}\right\|_2-\left\|g_{anc}-g_{neg}\right\|_2 + \delta\right]_+,
    \label{eq:triplet}
\end{equation}
where $g_{anc}, g_{pos}, g_{neg} \in \mathbb{R}^{256}$ are the global feature associated with anchor, positive and negative point clouds $\mathcal{P}_{anc}, \mathcal{P}_{pos}, \mathcal{P}_{neg}$, respectively, $\left[\cdot\right]_+$ denotes the hinge loss and $\delta$ is a margin hyperparameter.  In order to select our positive and negative point clouds we rely on the label $y_{a,b}$ which indicates if two point clouds $\mathcal{P}_a, \mathcal{P}_b$ are positive associated, negatively associated, or neither.  $y_{a,b}$ is generated using the ground truth sensor pose such that:
\begin{equation}
    \label{eq:source_labels}
    y_{a,b} = 
    \begin{cases}
        \mathrm{Positive} & s\left(t_a,t_b\right) \leq T_{pos} \\
        \mathrm{Negative} & s\left(t_a,t_b\right) \geq T_{neg} \\
        \mathrm{Neither} & \mathrm{Otherwise} \\
    \end{cases},
\end{equation}
where $s\left(t_a,t_b\right)$ denotes the distance between the ground truth sensor poses for $\mathcal{P}_a, \mathcal{P}_b$ and $T_{pos}, T_{neg}$ denote positive and negative distance thresholds in the world.  

For our local features $l\in \mathbb{R}^{\left|\mathcal{P}\right|\times 16}$, we follow \cite{vid2022logg3d} and use a combination of ground truth pose and ICP \cite{besl1992method} to align two positively associated point clouds $\mathcal{P}_a$, $\mathcal{P}_b$ and retrieve a set of point correspondences $C^{a\leftrightarrow b}$, where each correspondence is denoted as $c_i \in C^{a\leftrightarrow b} = \left\{\left(x_a^i,l_a^i\right), \left(x_b^i,l_b^i\right )\right\}$ with $x_a^i ,l_a^i , x_b^i ,l_b^i$ denoting the 3D point co-ordinate and local feature from $\mathcal{P}_a$,$\mathcal{P}_b$ linked by correspondence $c_i$.  The local features are then trained using the hardest contrastive loss:

\begin{equation}
    \begin{split}
        \mathcal{L}_{local} &= \sum_{c_i\in C^{a\leftrightarrow b}} \biggl\{ \left[\left\|l_a^i - l_b^i\right\|^2_2 - m_p\right]_+ / \left|C^{a\leftrightarrow b}\right| \\
        &+\lambda_n\left[m_n-\min_{k\in\mathcal{M}}\left\|l_a^i-l_b^k\right\|^2_2\right]_+ / \left|C^{a\leftrightarrow b}\right| \\
        &+\lambda_n\left[m_n-\min_{k\in\mathcal{M}}\left\|l_b^i-l_a^k\right\|^2_2\right]_+ / \left|C^{a\leftrightarrow b}\right| \biggl\},\\
    \end{split}
\end{equation}

where $\mathcal{M}$ is a random subset of features used for hard negative mining, hyperparameters $m_p, m_n$ are scalar margins and $\lambda_n$ is a scalar weight.  This gives us a combined training loss for the global and local features of:

\begin{equation}
    \mathcal{L}_{\mathrm{Total}} = \mathcal{L}_{triplet} + \mathcal{L}_{local},
\end{equation}
where $\mathcal{L}_{triplet}$ and $\mathcal{L}_{local}$ are the losses to train the local and global features respectively.

\subsection{Geometric Consistency Classifier}
\label{sec:GGPCM}
The second component of our auxiliary classification head is our proposed Geometric Consistency Classifier (GCC), which we train on top of the frozen local feature extractor outlined in the previous section.  {While local point features are no less susceptible to domain shift than global point cloud embeddings, by leveraging the geometric consistency of inlier correspondences as a prior our proposed GCC is able to produce reliable and accurate pseudo-labels for target adaptation in a manner which is robust against the deleterious impact of domain shift between the source and target datasets.}

Given two point clouds $\mathcal{P}_a$, $\mathcal{P}_b$ we extract local features $l_a $, $l_b$ before using a nearest neighbour search on the local features to produce a set of proposed point correspondences between the two point clouds $\hat{C}^{a\leftrightarrow b}$.  Next, we construct a geometric consistency matrix $M_{a,b}$ for which each entry $m_{i,j}$ measures the pairwise length consistency between correspondences $c_i, c_j$ in $\hat{C}^{a\leftrightarrow b}$, which is defined as:

\begin{equation}
    m_{i,j} = \left[1-\frac{d_{i,j}^2}{d_{thr}^2}\right]_+, d_{i,j} = \left|\left\|x_a^i-x_a^j\right\|_2 - \left\|x_b^i-x_b^j\right\|_2\right| , 
\end{equation}
where $d_{thr}^2$ is a hyperparameter which controls sensitivity to length difference.  As seen in Figure \ref{fig:spatial_consistency} when two point clouds are positively correlated $M_{a,b}$ is dominated by a large cluster of geometrically consistent inliers, while no such cluster exists for negatively associated point clouds.  Inspired by \cite{spectralmatch} we can consider the leading eigenvector $e_{a,b}\in\mathbb{R}^{\left|\hat{C}^{a\leftrightarrow b}\right|}$ of $M_{a,b}$ to be the association of each correspondence $c_i \in \hat{C}^{a\leftrightarrow b}$ with the main cluster of $M_{a,b}$, and therefore a good estimation for inlier probability.  We observe in Figure \ref{fig:spatial_consistency} that not only is the distribution of values in $e_{a,b}$ clearly distinct between positively and negatively associated point clouds, but also that the chances of two negatively associated point clouds forming a large spatially consistent cluster in order to mimic the distribution of a positive match is extraordinarily small.  Therefore, we feed the inlier probability into an MLP to produce a pair-likelihood score as follows:

\begin{equation}
    \beta_{a,b} = \sigma\left(f\left(\hat{e}_{a,b};\theta\right)\right),
\end{equation}
where $\beta_{a,b}$ is the pair-likelihood score for point clouds $\mathcal{P}_a$, $\mathcal{P}_b$, $f\left(\cdot;\theta\right)$ is an MLP parameterised by $\theta$, $\sigma$ is the sigmoid function, and $\hat{e}_{a,b}$ is ${e}_{a,b}$ with values sorted and scaled into the range $(0,1)$.   {Unlike common registration methods such as ICP \cite{besl1992method} the GCC does not require any initial alignment in order to reliably associate the input clouds, leading to more reliable detection of orthogonal or reverse revisits which provide valuable hard positive examples to the pseudo-label set.} To train the GCC we use the binary cross-entropy loss: 

\begin{equation}
    \mathcal{L}_{cls} = -\left(y_{a,b} \cdot \log \beta_{a,b} + \left(1-y_{a,b}\right)\log \left(1-\beta_{a,b}\right)\right),
\end{equation}

where $y_{a,b}$ is 1 for a positive pair and 0 otherwise.  {The GCC is trained on $\mathcal{S}$ where there is full access to the ground-truth supervision, but requires no re-training or ground truth when used for pseudo-label generation.}

\subsection{Pseudo-Label Generation}
Now with both the global and auxiliary classification heads fully trained on $\mathcal{S}$, we can generate pseudo-labels for adaptation to $\mathcal{T}$.  For each target point cloud $\mathcal{P}_a \in \mathcal{T}$ we extract global and local features $g_a,l_a$.  Next for each point cloud, we use the $L_2$ distance between global descriptors to retrieve the top-K nearest neighbours in the feature space as a list of potential positive and negative candidate point clouds, $\left\{\mathcal{P}_1,\mathcal{P}_2,\cdots,\mathcal{P}_K\right\}$.  We then generate pseudo-labels for each point cloud and its list of candidates as follows:

\begin{equation}
    y'_{a,b} = 
    \begin{cases}
        \mathrm{Positive} & \beta_{a,b} \geq \alpha_{pos} \\
        \mathrm{Negative} & \beta_{a,b} \leq \alpha_{neg} \\
        \mathrm{Neither} & \mathrm{Otherwise} \\
    \end{cases} , 
\end{equation}
where $\alpha_{pos}, \alpha_{neg}$ are scalar thresholds used to define whether the pair-likelihood score indicates a positive or negative example, and $b \in \left\{1,2,\cdots K\right\}$ is the index of the candidate point cloud.  The advantage of implementing these confidence thresholds is that underneath a severe domain shift, the pre-trained model features often may not be informative enough to reliably identify how correlated two input point clouds are.  The confidence thresholds allows us to identify when the GCC is uncertain about a pair of point clouds and exclude them from the set of pseudo-labels used for re-training, improving the reliability of our test-time adaptation.

\vspace{-2mm}
\subsection{Target Re-Training}
When adapting the model to the target dataset we discard the local feature decoder and GCC, re-training only the encoder and global feature decoder.  The model is fine-tuned (from source pre-trained weights) with Equation \ref{eq:triplet}, using the generated pseudo-labels $y'_{a,b}$ in place of ground-truth supervision.  If for a point cloud $\mathcal{P}_a \in \mathcal{T}$ either no positive or no negative examples were found in the previous step, then that point cloud is excluded as an anchor during re-training.

\section{Experimental Settings}

\begin{table}[t!]
    \centering
    \caption{{Details of source and target training and evaluation sets. * indicates shared training set for Wild-Places\cite{knights2023wildplaces}.}}
    \begin{NiceTabular}{l|c|c|c}
         Dataset & Setting & Sensor  & Number of Scans \\
        \hline 
         \textit{Source:} & & & (train) \\
         MulRan \cite{mulran20} & {Urban} & OS1-64 & 29,112 \\
         KITTI \cite{geiger2013kittidataset} & {Urban} & HDL-64E & 23,201 \\
         \hline 
         \textit{Target (Moderate):} & &  & (train / query / database) \\
         ALITA \cite{yin2022alita} & {Urban} & VLP-16 & 15,439 / 451 / 576\\
         NCLT \cite{ncarlevaris-2015a} & Campus & HDL-32E  & 20,503 / 4,486 / 26,945\\
         \hline 
        \textit{Target (Severe):} & & \\
        {Wild-Places (V)}\cite{knights2023wildplaces} & Natural & VLP-16  & 19,096* / 6,395 / 23,472\\
        {Wild-Places (K)}\cite{knights2023wildplaces}  & Natural & VLP-16 & 19,096* / 9,642 / 39,847\\

    \end{NiceTabular}
    
    \label{tab:datasets}
\end{table}

\label{sec:experiments}
\subsection{Training and Implementation Details}
We train our approach on 4 NVIDIA Tesla P100-16GB GPUs, and follow the data augmentation strategy proposed by \cite{Komorowski_2021_WACV}.  {We set the hyperparameters $\alpha_{pos}, \alpha_{neg}$ to 0.95 and 0.2 respectively for all experiments based on our ablations in Section \ref{sec:ablations}, and the number of considered candidates $K$ to 50.}  We use a sparse-convolutional architecture with skip connections for our encoder and decoders, and GeM \cite{gem19} pooling in our global feature head. For the source datasets we train the model from scratch for 80 epochs with a learning rate of {1e-3} decayed by a factor of 10 at 30 and 60 epochs, and the GCC is trained for 5 epochs using a learning rate of 0.01 decayed with a cosine scheduler.  When re-training we fine-tune the pre-trained model for 40 epochs at a learning rate of {1e-4}, decayed by a factor of 10 at 25 epochs.

\vspace{-2mm}

\subsection{Datasets}
\label{sec:datasets}
We evaluate our proposed approach using two source datasets and three target datasets.  \textbf{Source: } We use the MulRan \cite{mulran20} and KITTI Odometry \cite{geiger2013kittidataset} datasets for source pre-training, both of which were collected using vehicle-mounted sensors in urban environments.  We follow \cite{vid2022logg3d} and use a combination of MulRan sequences 01 and 02 from the DCC environment and sequences 01 and 03 from the Riverside environment for training, and use all  11 KITTI sequences when training.  \textbf{Target: } We further subdivide our target datasets into two categories: those experiencing a `moderate' domain shift, representing urban-to-urban {or urban-to-campus} TTA, and those experiencing a `severe' domain shift, representing the much more challenging urban-to-natural TTA scenario.  For the `moderate' datasets we use the ALITA \cite{yin2022alita} and NCLT \cite{ncarlevaris-2015a} datasets.  For ALITA we use the training, query and database splits proposed by the authors of the dataset, while for NCLT we use sequences (2012-01-08, 2012-01-15) for training and sequences (2012-11-04, 2012-11-16, 2012-11-17, 2012-12-01,  2013-02-23, 2013-04-05) for evaluation, following the same query/database split used in \cite{xu2021disco}.  For the `severe' target datasets we use the recently introduced Wild-Places \cite{knights2023wildplaces} benchmark dataset, a natural place recognition dataset recorded across two large-scale natural environments.  We use {Wild-Places (V)} and Wild-Places (K) to refer to the \textit{\underline{V}enman} and \textit{\underline{K}arawatha} environments in the Wild-Places dataset respectively, and use the \textit{inter-sequence} training and evaluation setup established by the authors. Table \ref{tab:datasets} outlines the environment, sensors and training splits for each of the source and target datasets used in our experiments.

\vspace{-2mm}

\subsection{Evaluation Metrics}
For evaluation, we calculate the $L_2$ distance between the global embedding for a query point cloud and point clouds from a database consisting of different traversals of the same region from the database.  We report mean Recall@N (R@N) for N = 1, 5, 1\%, considering a query point cloud to be successfully relocalised if one of the top-N retrieved database candidates is within 5m for ALITA \cite{yin2022alita} and within 3m for all other datasets.  We use SOURCE$\rightarrow$TARGET to denote the source and target datasets for a given result.

\section{Results}
\label{sec:results}
\begin{table*}[btp]
    \centering
    \caption{Test-Time Adaptation performance for LiDAR place recognition under ``moderate" domain shifts.  Rows above and below the dashed line represent source-only and target-adapted approaches respectively.  Bold and underlined values are first and second place respectively for a given column.}
    \begin{NiceTabular}{wl{2.7cm}||wc{0.56cm}wc{0.56cm}wc{0.56cm}|wc{0.56cm}wc{0.56cm}wc{0.56cm}||wc{0.63cm}wc{0.63cm}wc{0.63cm}|wc{0.63cm}wc{0.63cm}wc{0.63cm}||wc{0.44cm}wc{0.44cm}wc{0.44cm}}
        \hline 
        \Block{2-1}{Method} & \Block{1-3}{KITTI$\rightarrow$ALITA} & & & \Block{1-3}{KITTI$\rightarrow$NCLT} & & & \Block{1-3}{MulRan$\rightarrow$ALITA} & & & \Block{1-3}{MulRan$\rightarrow$NCLT} & & & \Block{1-3}{Average} \\
        & R@1 & R@5 & R@1\% & R@1 & R@5 & R@1\% & R@1 & R@5 & R@1\% & R@1 & R@5 & R@1\% & R@1 & R@5 & R@1\%\\
        \hline 
        ScanContext~\cite{kim2018scan} & 56.03 & 77.07 & 59.59 & {64.74} & {69.54} & {74.88} & 56.03 & 77.07 & 59.59 & 64.74 & 69.54 & 74.88 & 60.39 & 73.31 & 67.24 \\
        MinkLoc3Dv2~\cite{komorowski2022improving} & 72.74 & 92.75 & 77.44 & 54.58 & 71.38 & 84.42 &  70.21 & 91.05 & 75.08 & {69.01} & {86.39} & \textbf{95.70} & 66.64 & 85.39 & 83.16 \\
        LoGG3D-Net~\cite{vid2022logg3d} & 59.68 & 82.76 & 64.72 & 41.80 & 58.43 & 73.37 & 71.19 & 92.57 & 75.26 & 33.28 & 47.12 & 58.95 & 52.02 & 70.41 & 68.98 \\
        \hdashline
        LoGG3D-Net+SGV~\cite{vidanapathirana2023sgv} & 88.32 & 91.97 & 88.64 &57.82&67.36&73.37& 93.24 & 96.83 & 93.9 &50.48&56.48&58.95&72.47&78.16&78.72\\
        \textit{\textit{GeoAdapt}} & \textbf{96.95} & \textbf{100.0} & \textbf{98.25} & \underline{69.45} & \underline{85.55} & \underline{\textbf{94.63}} & \textbf{97.56} & \textbf{100.0} & \textbf{98.69} & \underline{71.98} & \underline{88.03} & {\underline{95.68}} & \underline{83.99} & \underline{93.40} & \textbf{96.81}\\
        \textit{\textit{GeoAdapt}}+SGV & \underline{96.91} & \underline{100.0} & \underline{98.22} &\textbf{82.83}&\textbf{91.41}&\underline{\textbf{94.63}}&\underline{97.13} & \underline{99.83} & \underline{98.60}&\textbf{83.82}&\textbf{91.86}&{\underline{95.68}}&\textbf{90.17}&\textbf{95.78}&\underline{96.78} \\
        \hline 
        
    \end{NiceTabular}
    
    \label{tab:soaeasy}
\end{table*}

\begin{table*}[btp]
    \centering
    \caption{{Test-Time Adaptation performance for LiDAR place recognition under ``severe" domain shifts.}}
    \begin{NiceTabular}{wl{2.7cm}||wc{0.56cm}wc{0.56cm}wc{0.56cm}|wc{0.56cm}wc{0.56cm}wc{0.56cm}||wc{0.63cm}wc{0.63cm}wc{0.63cm}|wc{0.63cm}wc{0.63cm}wc{0.63cm}||wc{0.44cm}wc{0.44cm}wc{0.44cm}}
        \hline
        \Block{2-1}{Method} & \Block{1-3}{KITTI$\rightarrow${Wild-Places(V)}} & & & \Block{1-3}{KITTI$\rightarrow${Wild-Places(K)}} & & & \Block{1-3}{MulRan$\rightarrow${Wild-Places(V)}} & & & \Block{1-3}{MulRan$\rightarrow${Wild-Places(K)}} & & & \Block{1-3}{Average} \\
        & R@1 & R@5 & R@1\% & R@1 & R@5 & R@1\% & R@1 & R@5 & R@1\% & R@1 & R@5 & R@1\% & R@1 & R@5 & R@1\%\\
        \hline 
        ScanContext~\cite{kim2018scan} & 33.98 & 48.79 & 61.56 & 38.44 & 53.56 & 66.16 & 33.98 & 48.79 & 61.56 & 38.44 & 53.56 & 66.16 & 36.21 & 51.18 & 63.86\\
        MinkLoc3Dv2~\cite{komorowski2022improving} &  5.40 & 17.15 & 41.82 &  6.49 & 18.79 & 42.53 & 11.18 & 26.47 & 50.01 & 14.35 & 33.32 & 58.98 & 9.36 & 23.93 & 48.36\\
        LoGG3D-Net~\cite{vid2022logg3d} &  6.00 & 15.13 & 32.54 &  8.55 & 19.43 & 37.26 &  3.91 & 11.69 & 27.97 &  5.16 & 14.07 & 30.43 & 5.91 & 15.08 & 32.05\\
        \hdashline
        LoGG3D-Net+SGV~\cite{vidanapathirana2023sgv} &18.84&23.89&32.54&21.85&28.34&37.26&14.8&20.94&27.97&15.56&22.8&30.43&17.76&23.99&32.05\\
        \textit{\textit{GeoAdapt}}     & \underline{60.48} & \underline{84.93} & \textbf{\underline{96.30}} & \underline{47.95} & \underline{75.23} & \textbf{\underline{93.02}} & \underline{51.99} & \underline{77.26} & \textbf{\underline{91.82}} & \underline{41.45} & \underline{69.66} & \textbf{\underline{89.54}} & \underline{50.56} & \underline{76.77} & \textbf{\underline{92.67}} \\
        
        \textit{\textit{GeoAdapt}}+SGV  & \textbf{74.12} & \textbf{90.74} & \textbf{\underline{96.30}} &\textbf{58.68}&\textbf{81.09}&\textbf{\underline{93.02}}&\textbf{70.34}&\textbf{86.96}&\textbf{\underline{91.82}}&\textbf{56.39}&\textbf{78.04}&\textbf{\underline{89.54}}&\textbf{64.88}&\textbf{84.21}& \textbf{\underline{92.67}} \\
        \hline
        
    \end{NiceTabular}
    \vspace{-5mm}
    \label{tab:soahard}
\end{table*}

\subsection{Comparison to State-of-the-Art}
Tables \ref{tab:soaeasy} and \ref{tab:soahard} compare the performance of \textit{\textit{GeoAdapt}} to existing state-of-the-art approaches for place recognition for `moderate' and `severe' domain shifts respectively.  We compare against one handcrafted approach, ScanContext \cite{kim2018scan}, and two state-of-the-art learning-based approaches, MinkLoc3Dv2 \cite{komorowski2022improving} and LoGG3D-Net \cite{vid2022logg3d}.  {As previously mentioned TTA approaches for tasks such as segmentation or object detection are generally not applicable to LiDAR place recognition due to fundamental differences in the ground truth required for training, and to the best of our knowledge no other approach for Test-Time Adaptation on LiDAR place recognition exists in the literature.  Therefore we also compare against SGV \cite{vidanapathirana2023sgv}, which performs re-ranking on the target to improve performance at test time without requiring re-training or access to target supervision.}  %

We observe that the exclusively source pre-trained methods report significantly diminished results when evaluated on out-of-domain data, achieving a maximum R@1 of only 66.64\% and 9.36\% on the `moderate' and `severe' target datasets respectively. By adapting the model to the target data distribution \textit{\textit{GeoAdapt}} is able to maintain high performance on the target datasets without having required access to any ground truth information during adaptation, reporting a R@1 performance of 83.99\% (+17.35\%) and 50.56\% (+41.2\%) on the `moderate' and `severe' datasets.  {These results illustrate the impact that the domain shifts induced by a change in environment and LiDAR sensor can have on model generalisability, and the effectiveness of \textit{\textit{GeoAdapt}} in addressing this challenge.}

Comparing to SGV \cite{vidanapathirana2023sgv}, we observe that when applied to LoGG3D-Net \cite{vid2022logg3d} SGV improves  performance by 20.45\% and 11.85\% on the `moderate' and `severe' target datasets respectively but is still out-performed by \textit{\textit{GeoAdapt}}.  The upper bound of performance for re-ranking methods is limited by the quality of the proposed re-ranking candidates, which can be an issue when - as particularly apparent in Table \ref{tab:soahard} - domain shifts severely degrade the quality of the initial proposals.  We observed that the highest performance overall comes from combining \textit{\textit{GeoAdapt}} with SGV, ``bootstrapping" the model on the target domain to improve the quality of the re-ranking candidates and boosting the performance of \textit{\textit{GeoAdapt}} by an additional 6.18\% and 14.32\% on the `moderate' and `severe' target datasets respectively.

\begin{figure}[t]
\begin{subfigure}[b]{0.49\columnwidth}
         \centering
         \includegraphics[width=\textwidth]{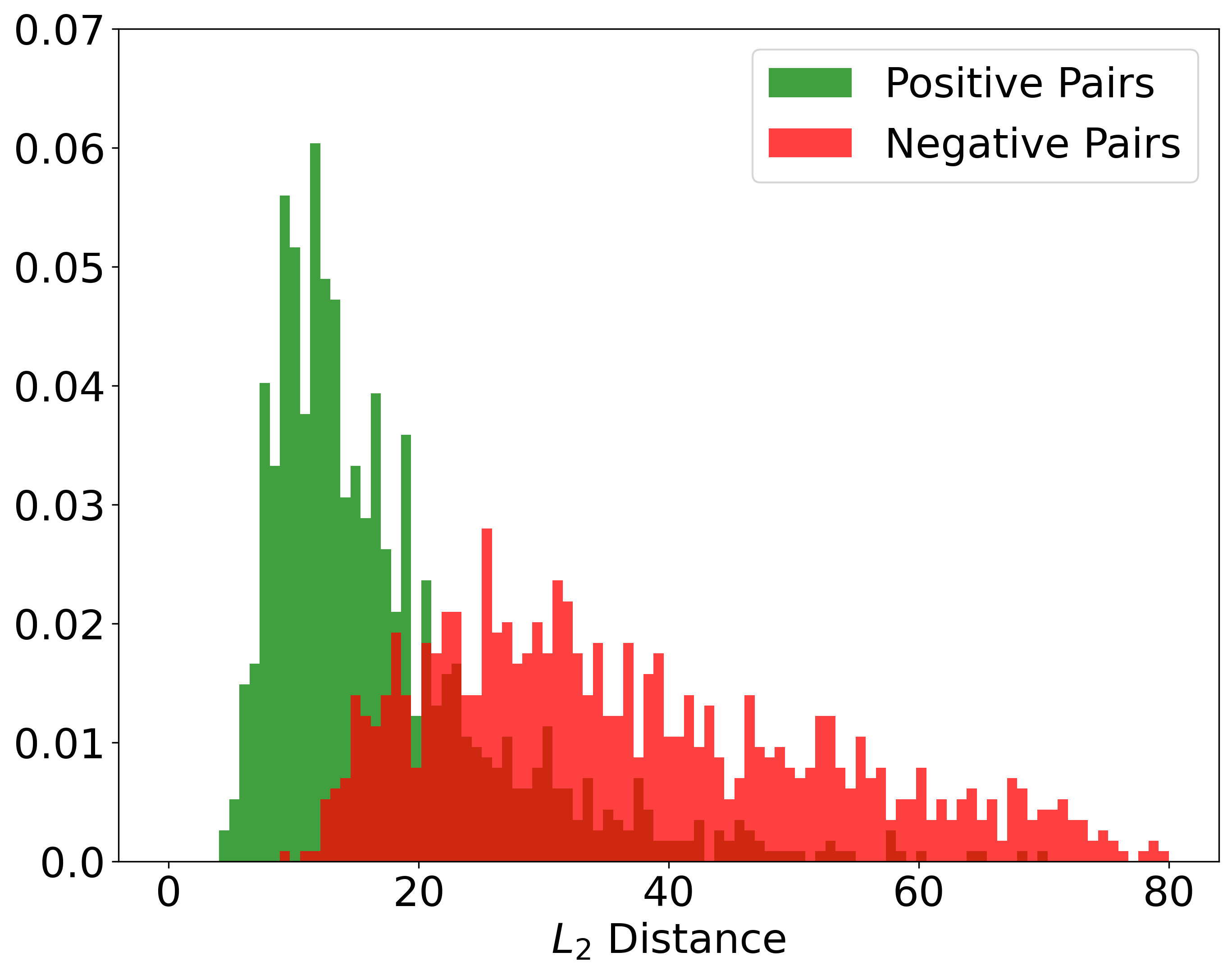}
         \caption{Before \textit{\textit{GeoAdapt}}}
     \end{subfigure}
     \hfill 
    \begin{subfigure}[b]{0.49\columnwidth}
         \centering
         \includegraphics[width=\textwidth]{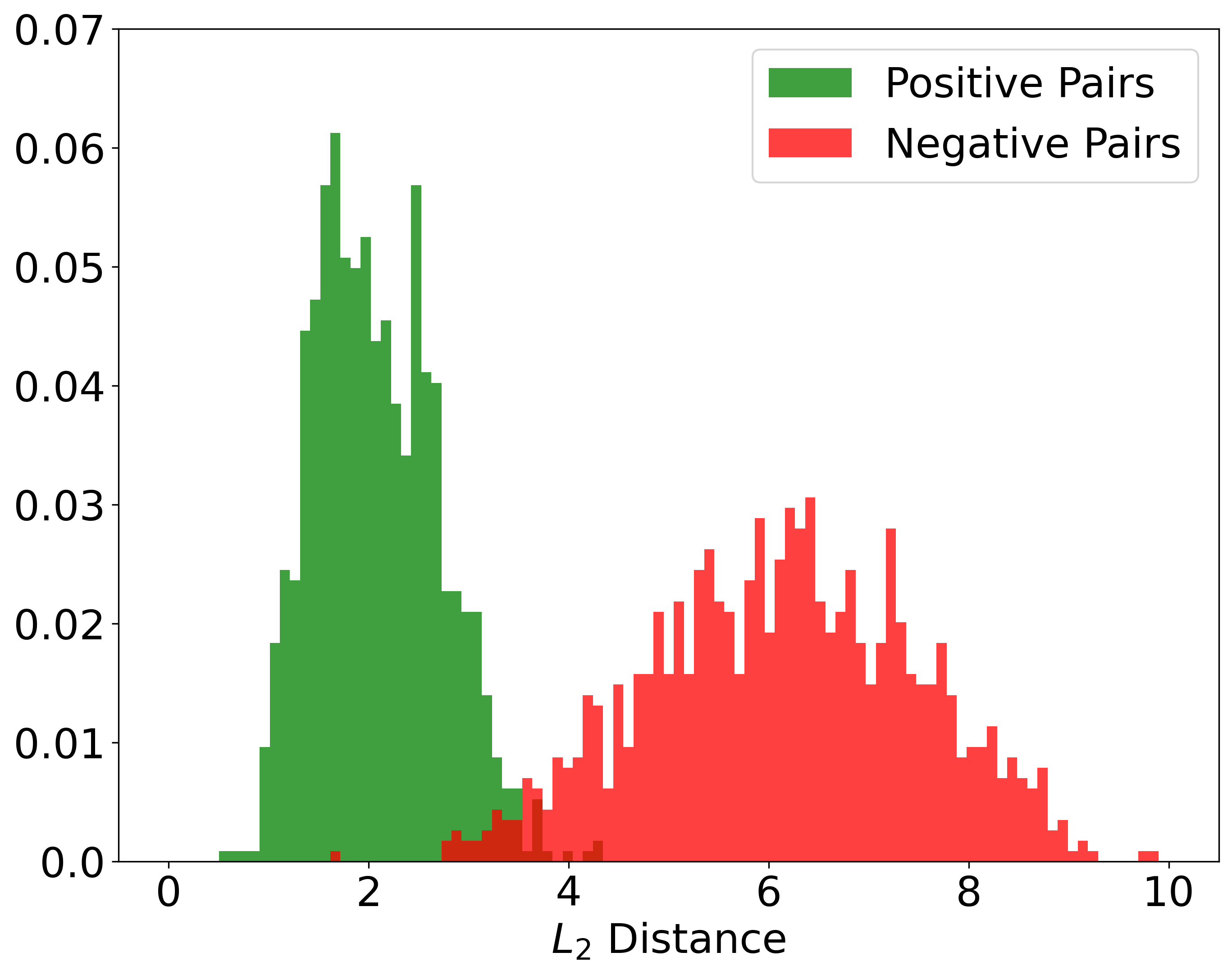}
         \caption{After \textit{\textit{GeoAdapt}}}
     \end{subfigure}

    \caption{{Comparison of $L_2$ distance between positive and negative pairs of global embeddings for the source pre-trained model (a) and \textit{GeoAdapt} (b).  \textit{GeoAdapt} significantly improves contrast between positive and negative pairs on the target domain after re-training, leading to more reliable revisit detection and retrieval.   Results reported on KITTI$\rightarrow${Wild-Places (K)}.}}
    
    \label{fig:l2_dist_histogram}
    \vspace{-5mm}
\end{figure}

\begin{figure}[t]
\begin{subfigure}[b]{0.49\columnwidth}
         \centering
         \includegraphics[width=\textwidth]{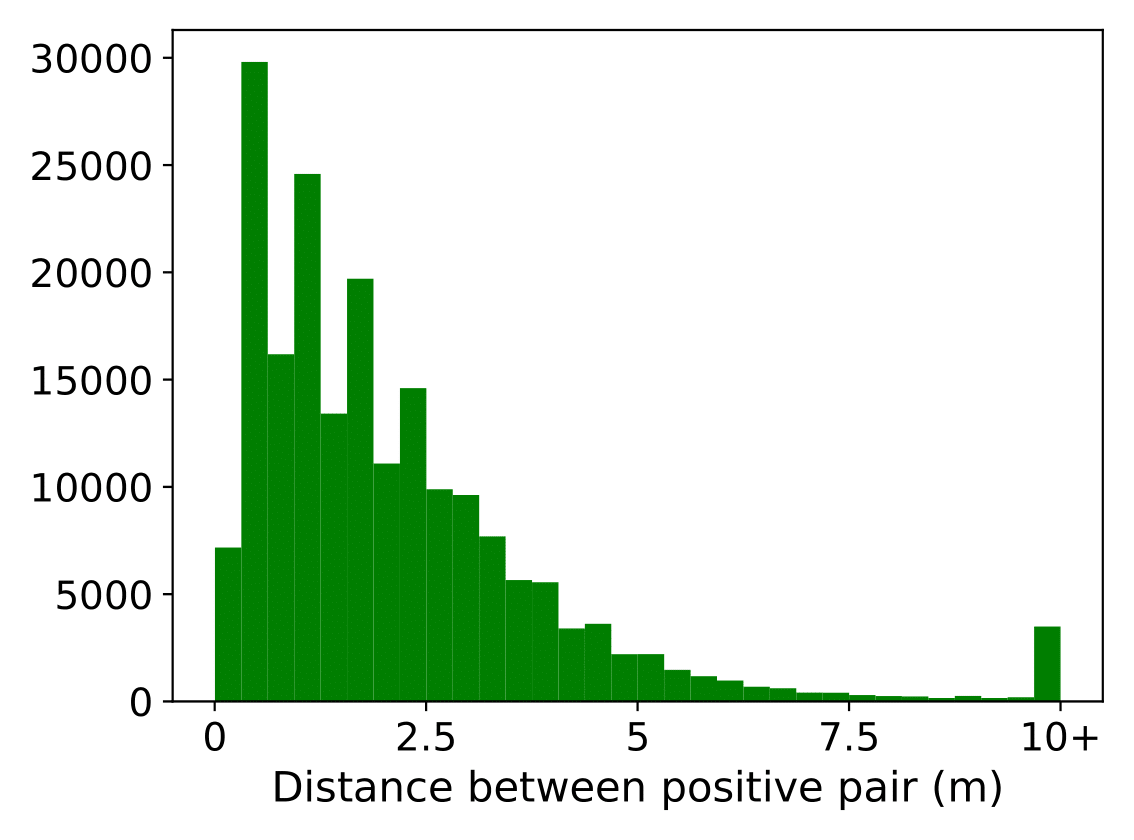}
         \caption{KITTI$\rightarrow$NCLT}
     \end{subfigure}
     \hfill 
    \begin{subfigure}[b]{0.49\columnwidth}
         \centering
         \includegraphics[width=\textwidth]{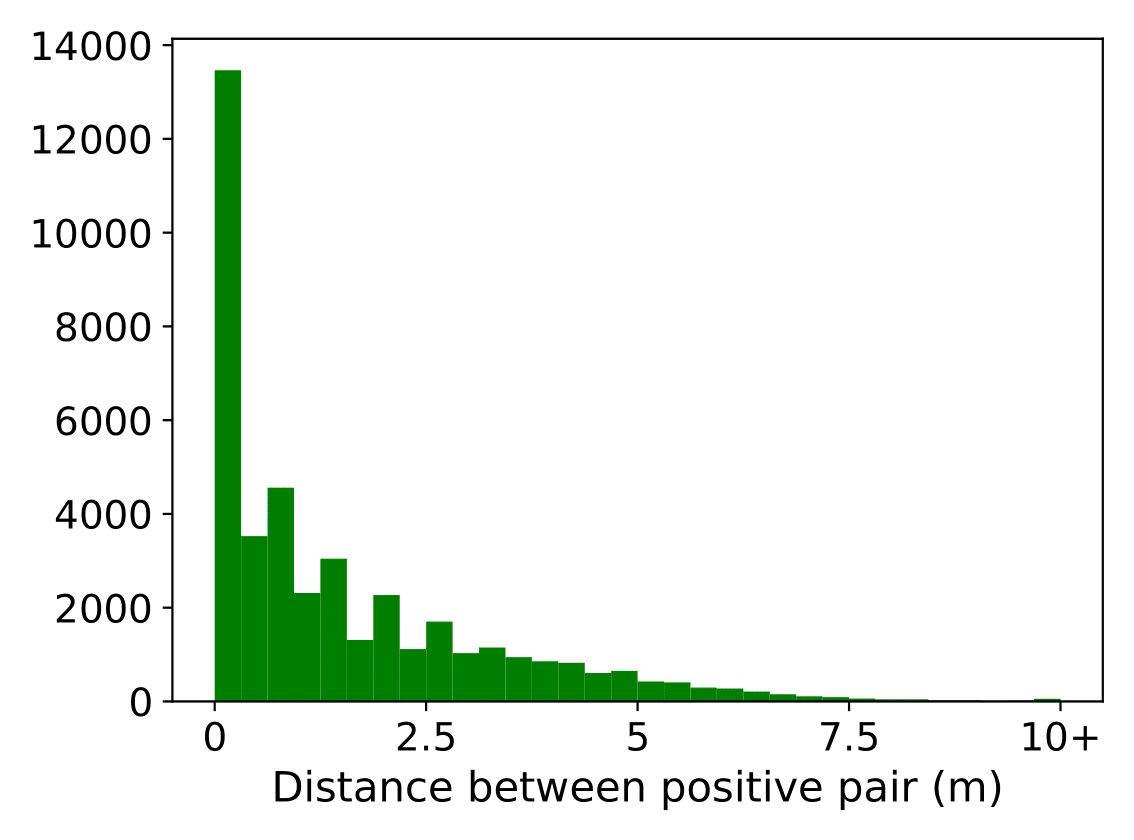}
         \caption{KITTI$\rightarrow${Wild-Places (K)}}
     \end{subfigure}

    \caption{%
    Histogram of distances between positive pairs of point clouds selected by \textit{GeoAdapt}. \textit{GeoAdapt} generates pseudo-labels based on geometric consistency rather than a scalar distance threshold, resulting in a more flexible positive selection for target re-training.
}
    \label{fig:pos_neg_histogram}
    \vspace{-5mm}
\end{figure}

\begin{figure*}[t]
    \centering
    \begin{subfigure}[b]{0.23\textwidth}
         \centering
         \includegraphics[width=\textwidth]{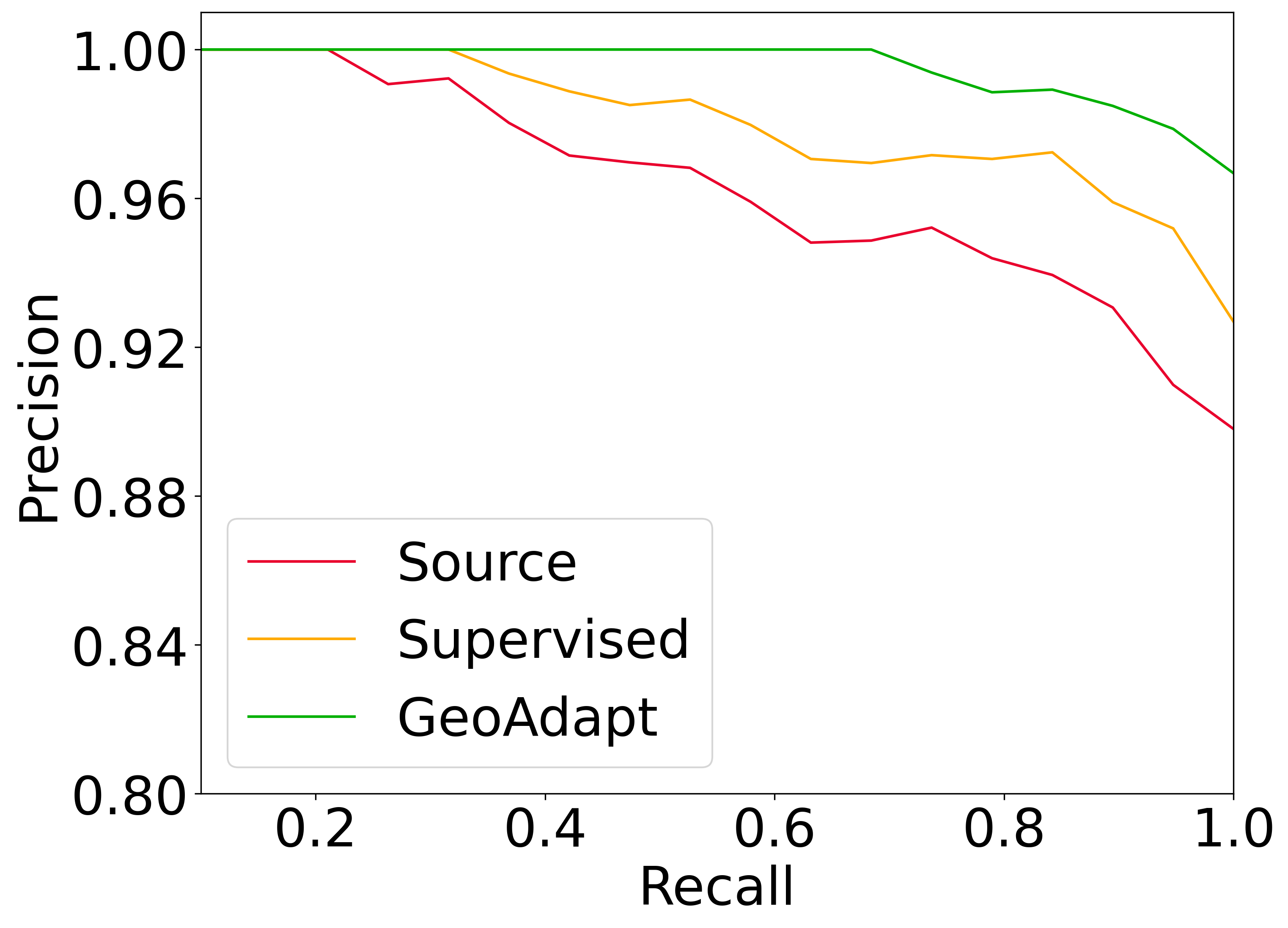}
         \caption{KITTI$\rightarrow$ALITA} 
         \label{fig:PR_ALITA}
     \end{subfigure}
     \hfill
     \begin{subfigure}[b]{0.23\textwidth}
         \centering
         \includegraphics[width=\textwidth]{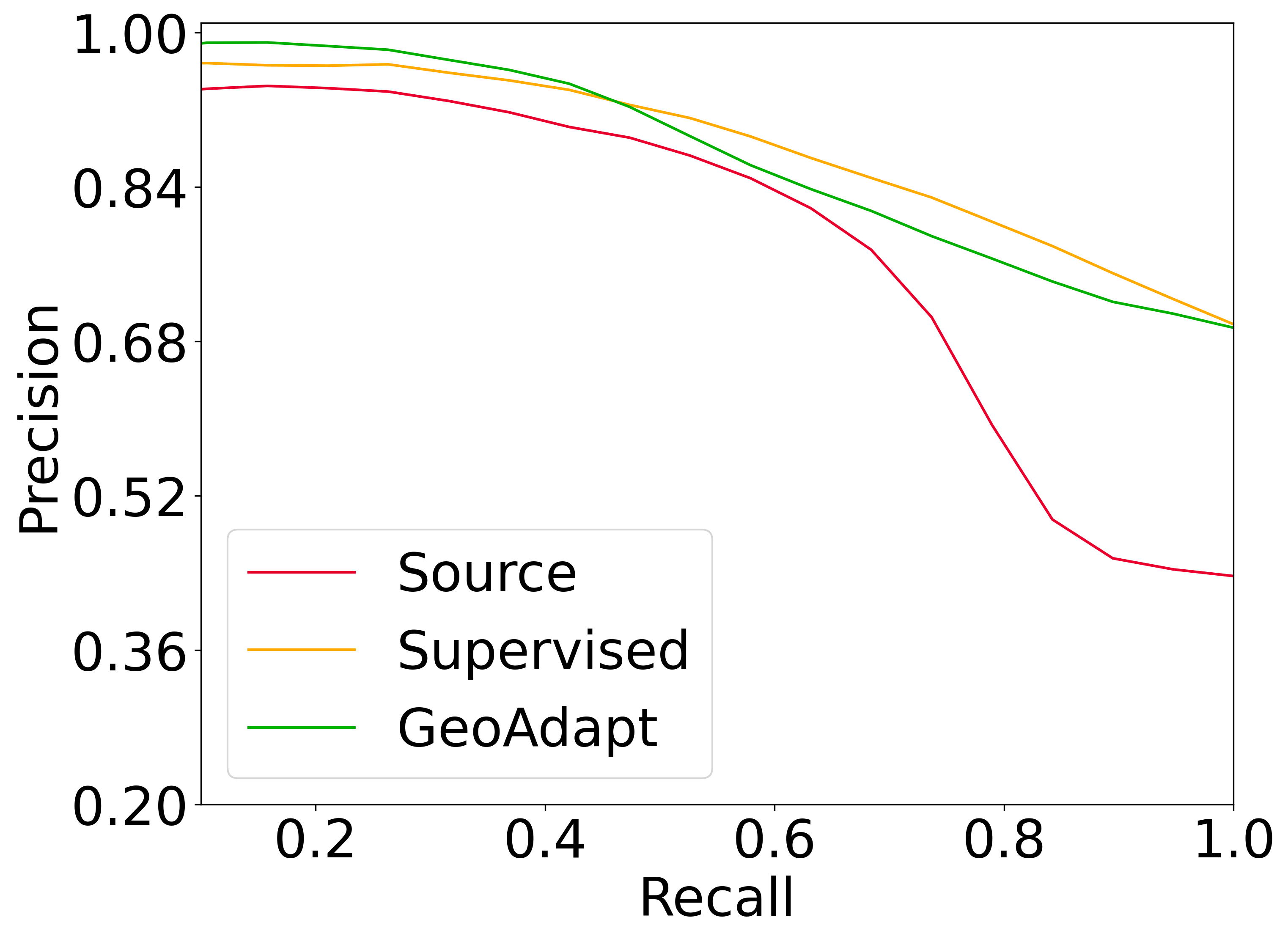}
         \caption{KITTI$\rightarrow$NCLT} 
         \label{fig:PR_NCLT}
     \end{subfigure}
     \hfill
     \begin{subfigure}[b]{0.23\textwidth}
         \centering
         \includegraphics[width=\textwidth]{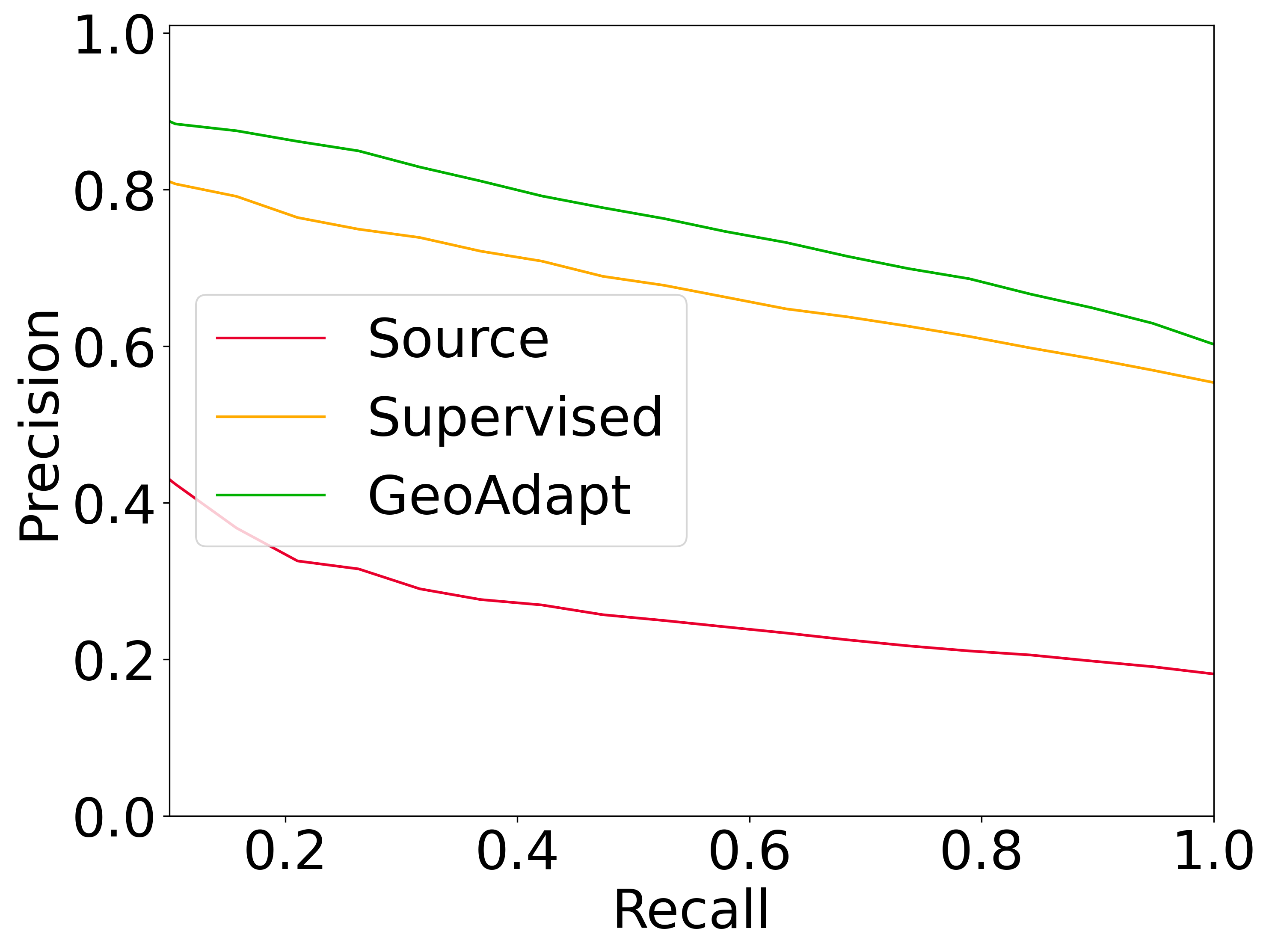}
         \caption{KITTI$\rightarrow${Wild-Places (V)}} 
         \label{fig:PR_Venman}
     \end{subfigure}
     \hfill
     \begin{subfigure}[b]{0.23\textwidth}
         \centering
         \includegraphics[width=\textwidth]{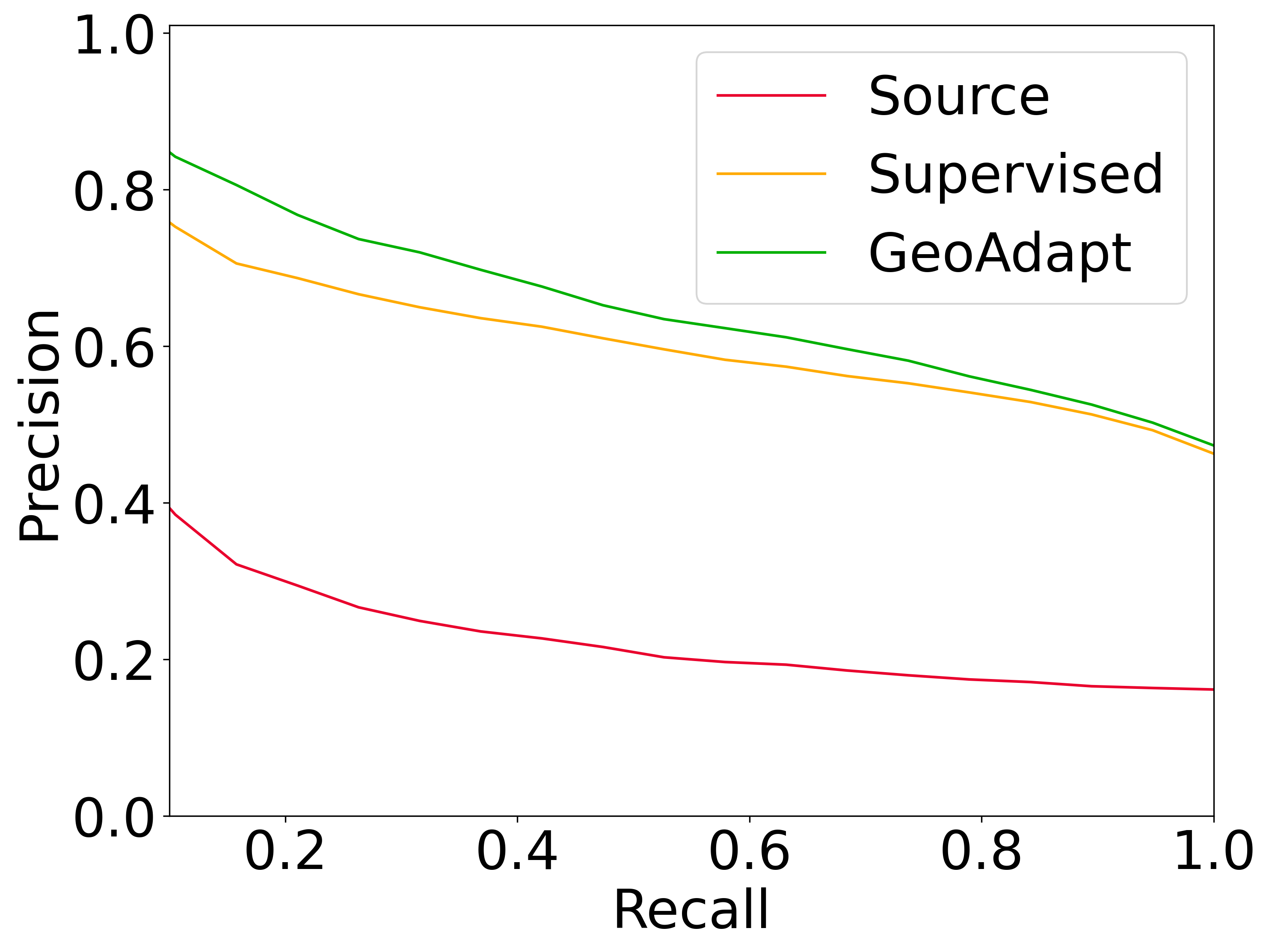}
         \caption{KITTI$\rightarrow${Wild-Places (K)}} 
         \label{fig:PR_Karawatha}
     \end{subfigure}
     
    \caption{Precision-Recall curves for models pre-trained on source data, re-trained with supervised ground truth, and re-trained with \textit{\textit{GeoAdapt}}.  \textit{\textit{GeoAdapt}} outperforms not only the source pre-training but also in some cases ground truth supervision. }
    \vspace{-5mm}
    \label{fig:prec_recall}
\end{figure*}

\vspace{-2mm}

\subsection{Comparison to Pre-Trained and Fully Supervised Models}
For place recognition, the model's feature space should demonstrate strong separability between global embeddings which are positively and negatively associated.  Domain shifts between the source and target data can severely degrade this separability, which in turn degrades our ability to use the global features to identify and retrieve place re-visits at evaluation.  In Figure \ref{fig:l2_dist_histogram} we show that re-training with \textit{\textit{GeoAdapt}} significantly improves the separability of positive and negative examples on the target, leading to more reliable revisit detection and retrieval.  Figure \ref{fig:prec_recall} presents precision-recall curves for a source pre-trained model, a model re-trained with \textit{GeoAdapt}, and a model re-trained with the supervised ground truth on the target.  We observe that \textit{\textit{GeoAdapt}} not only clearly outperforms the source pre-trained model, but also performs competitively with and sometimes out-performs the model re-trained using the supervised ground truth.

\begin{figure}[t]
    \centering
    \begin{subfigure}[b]{\columnwidth}
         \centering
         \includegraphics[width=0.7\textwidth]{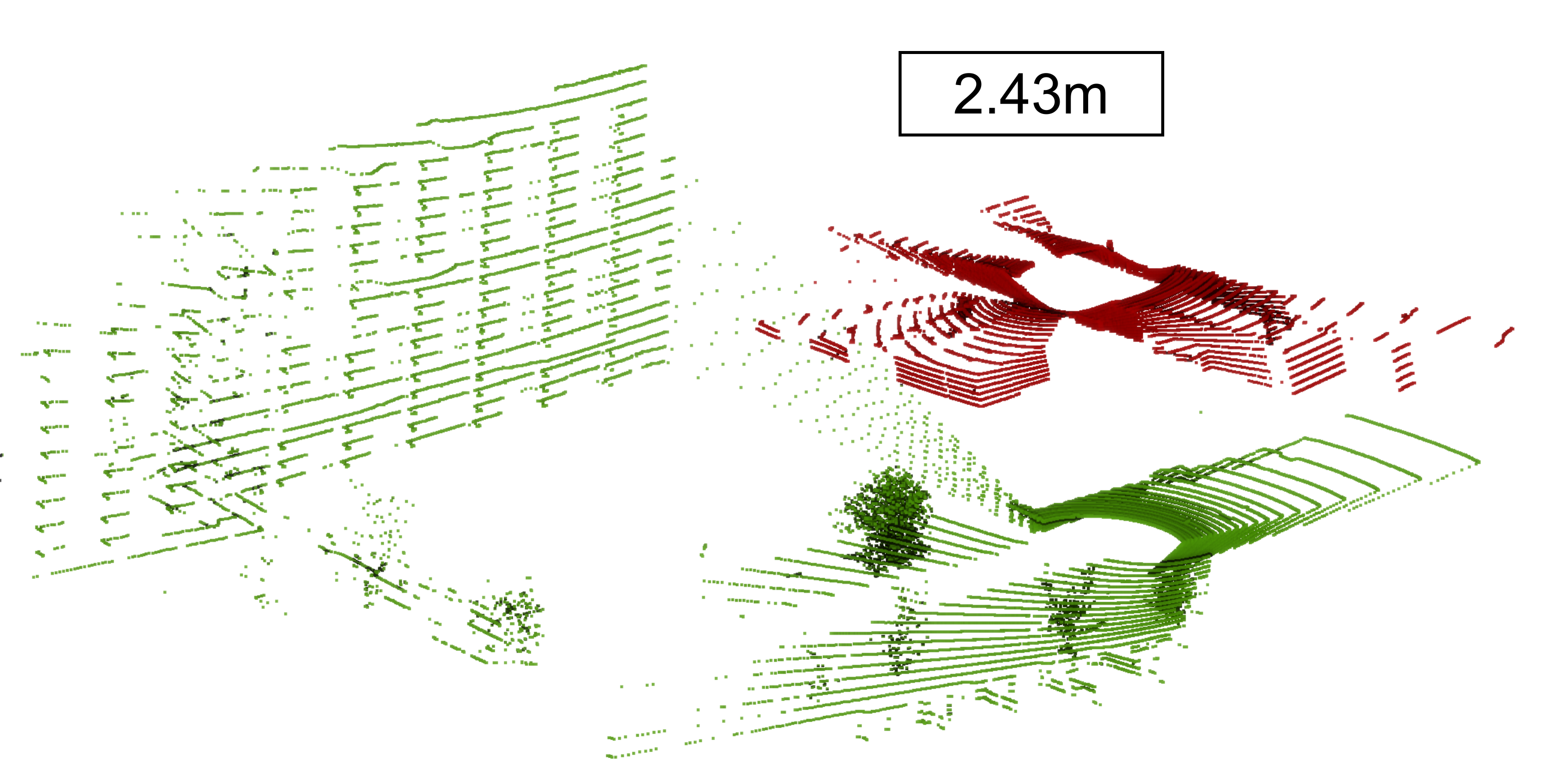}
         \caption{Positive pair selected by supervised ground truth}
         \label{fig:false_pos}
     \end{subfigure}
     
    \begin{subfigure}[b]{\columnwidth}
         \centering
         \includegraphics[width=0.7\textwidth]{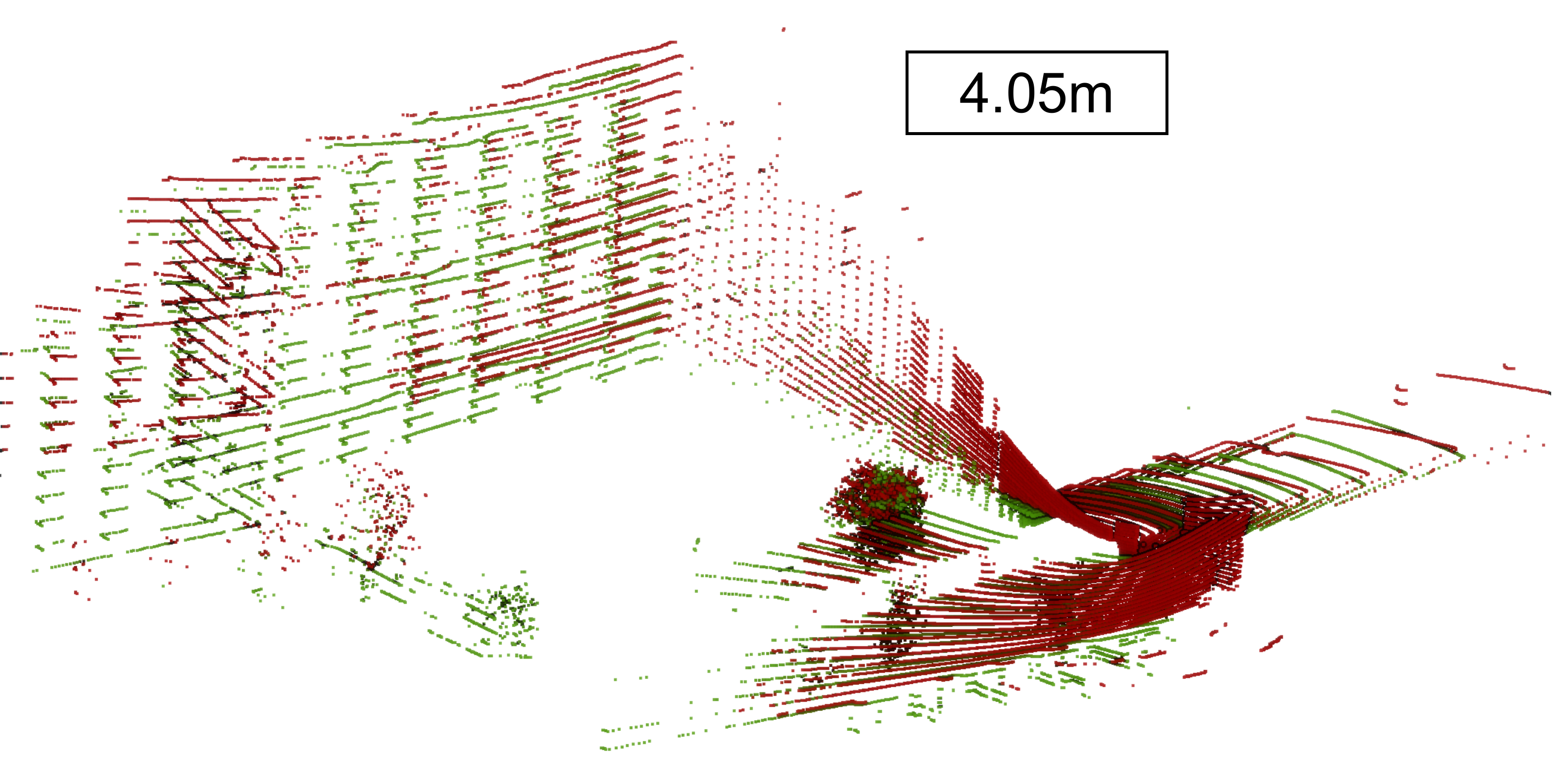}
         \caption{Positive pair selected by \textit{GeoAdapt}}
         \label{fig:our_pos}
     \end{subfigure}
     \caption{Examples of positive examples for supervised re-training and \textit{GeoAdapt}.  \textit{GeoAdapt} removes faulty positive pairs included by the supervised ground truth, and includes positives the supervised ground truth would otherwise exclude.}
     \vspace{-5mm}
    \label{fig:mismatch}
    
\end{figure}

We present several explanations for this behaviour.  Firstly, as has been raised in previous works \cite{Brachmann2021OnTL} the ground truth for place recognition datasets often contain errors due to noise in the GPS or SLAM-derived odometry ground truth, leading to faulty selection of positive and negative pairs which detrimentally impact the model's performance when used for re-training.  Secondly, as shown in Figure \ref{fig:mismatch} using a hard scalar threshold to form pairs for training presents some drawbacks even assuming a perfect ground truth pose.  Occlusion or other environmental factors can result in `positive' pairs of nearby point clouds which have little to no shared information (Figure \ref{fig:false_pos}), and highly correlated point clouds separated by a distance just slightly higher than the scalar threshold can be excluded from training despite having a great deal of shared information (Figure \ref{fig:our_pos}).

Comparatively, in Figure \ref{fig:pos_neg_histogram} we show that the positives selected by \textit{\textit{GeoAdapt}} follow a dynamic threshold guided by the geometric similarity of the inputs, which is effective in tackling the edge cases shown in Figure \ref{fig:mismatch}.  Our results suggest that the flexibility provided by using a learned approach to guide positive and negative selection can give an advantage over using an inflexible threshold on the target's supervised ground truth, though we leave further investigation of this phenomena for future work.

\vspace{-2mm}

\subsection{Ablation Studies}
\label{sec:ablations}
\begin{figure}[t]
    \centering
    \includegraphics[width=0.5\columnwidth]{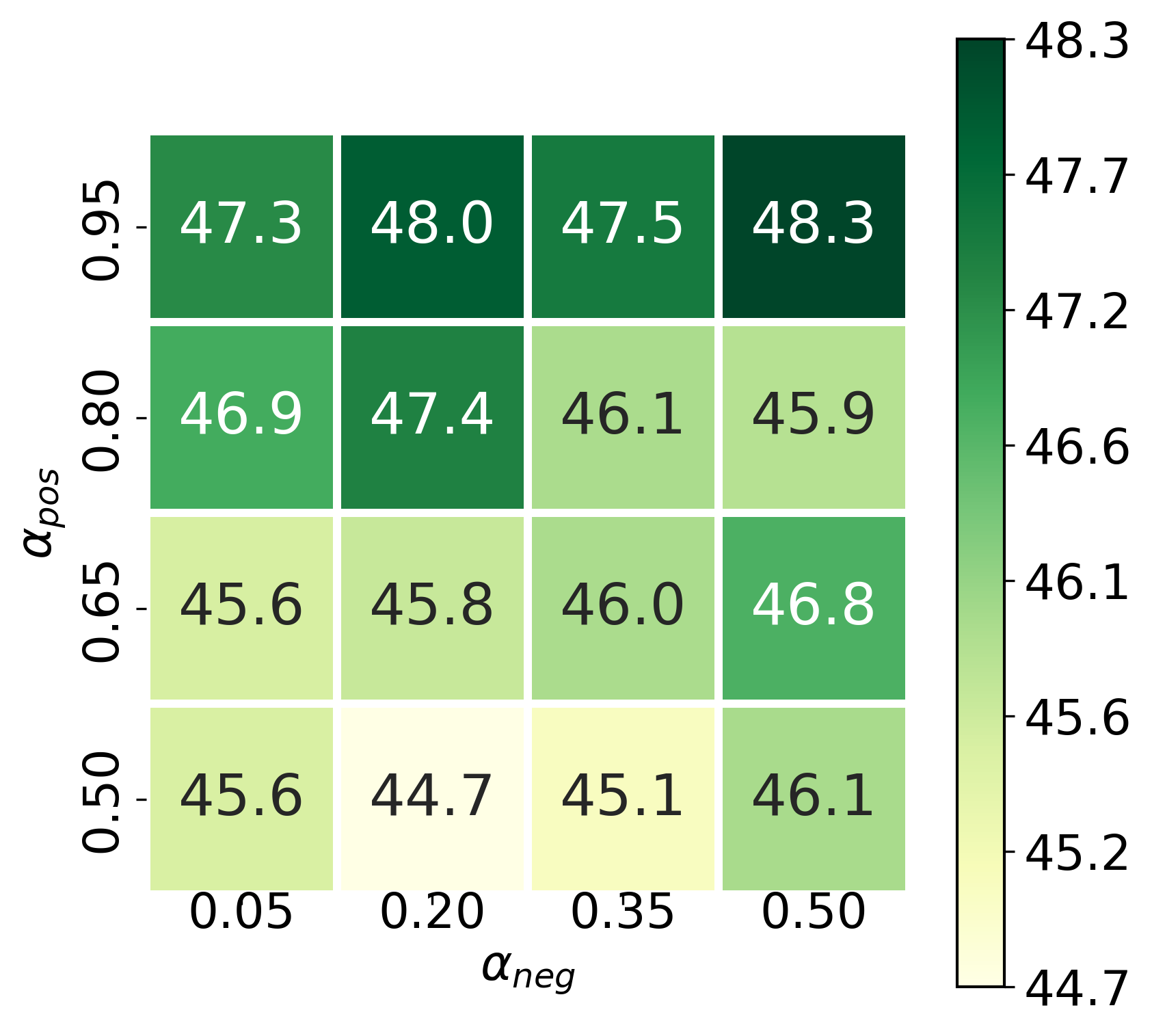}
    \caption{Impact of $\alpha_{pos}$, $\alpha_{neg}$ (Recall@1) on   KITTI$\rightarrow${Wild-Places (K)}.}
    \label{fig:thresh_ablation}
    \vspace{-5mm}
\end{figure}

In this section we explore the impact of how certain design choices and hyperparameter selection impacts the performance of the proposed approach.   Table \ref{tab:norm_sort_ablation} looks at the impact of pre-processing the pair-likelihood vector $e_{a,b}$ before its use as input to the MLP.  We observe that both sorting and scaling of $e_{a,b}$ are critical to the performance of the GCC, with Recall@1 performance dropping 23.54\% on the KITTI$\rightarrow${Wild-Places (K)} setup in their absence.  Figure \ref{fig:thresh_ablation} explores the impact of changing the value of hyperparameters $\alpha_{pos}, \alpha_{neg}$ on the KITTI$\rightarrow${Wild-Places (K)} scenario.  We observe that while employing a stricter value of $\alpha_{pos}$ has a notably positive impact on adaptation performance, no strong trend is present for changing the value of $\alpha_{neg}$. This result suggests that false positive examples have a significantly more deleterious impact on adaptation than false negatives, emphasising the importance of filtering them out of the set of pseudo-labels for the target data.  {We found that values for $\alpha_{pos}$ and $\alpha_{neg}$ selected through this ablation produce consistently strong results across all experimental setups as demonstrated by the results in Tables \ref{tab:soaeasy} and \ref{tab:soahard}, such that choosing new thresholds for new target datasets is not necessary.}

\begin{table}[t]
    \centering
    \caption{{Ablation study on the effects of scaling and sorting the inlier confidence scores.  Results are reported on KITTI$\rightarrow${Wild-Places (K)}.}}
    \begin{tabular}{wc{1.5cm}wc{1cm}|wc{0.5cm}wc{0.5cm}wc{0.5cm}}
        \hline 
         Scaling & Sorting & R@1 & R@5 & R@1\% \\
        \hline 
           - & -  & 24.41 & 49.14 & 72.21 \\ 
          \cmark & -  & 38.36 & 66.32 & 88.29 \\
          - & \cmark & 45.19 & 72.68 & 92.2 \\
          \cmark & \cmark & \textbf{47.95} & \textbf{75.23} & \textbf{93.02} \\
        \hline 
    \end{tabular}
    
    \label{tab:norm_sort_ablation}
\end{table}

\section{Conclusion}
\label{sec:conclusion}
In this paper we address the task of Test-Time Adaptation for LiDAR place recognition, adapting a source pre-trained model in a self-supervised manner to improve performance on an unlabelled target environment.  We propose a novel approach \textit{GeoAdapt}, which re-trains the model using robust pseudo-labels generated by an auxiliary classification head. 
 We demonstrate that geometric consistency as a prior when generating pseudo-labels results in strong and reliable adaptation to target environments underneath even severe domain shifts, and even performs competitively with fully supervised re-training on the target domain.  The ability of \textit{GeoAdapt} to outperform supervised re-training highlights several shortcomings in how ground truth is currently generated for place recognition datasets, and raises interesting avenues for future work investigating how geometric consistency can be used as a prior to guide both supervised and self-supervised learning in LiDAR place recognition.  We believe that this work opens several avenues for future exploration of TTA for localisation tasks.  These avenues include targeting specific domain shifts induced by factors such as changing sensor or weather effects, investigating how \textit{GeoAdapt} could be used to perform large-scale data annotation for city or larger scale place recognition, extending the work to TTA for 6-DoF metric localisation and addressing the additional challenges posed by self-supervised adaptation in the continual \cite{knights2022incloud} or online \cite{Vodisch2022ContinualSB} settings.

\section*{Acknowledgements}
This work was partially funded by CSIRO's Machine Learning and Artificial Intelligence Future Science Platform (MLAI FSP). The work was supported in part by an Australian Research Council (ARC) Discovery Program Grant No: DP200101942.

\balance{}

\bibliographystyle{./bibliography/IEEEtran}
\bibliography{ref}

\end{document}